\documentclass[preprint, 3p, twocolumn]{elsarticle}
\usepackage{graphicx}
\usepackage{xfrac}
\usepackage{amsmath}
\usepackage{amssymb}
\usepackage{dsfont}
\usepackage{parskip}
\usepackage{siunitx}
\usepackage{microtype}

\usepackage{url}

\begin{document}

\title{Sim-to-real transfer of active suspension control using deep reinforcement learning}
\author[1,2]{Viktor Wiberg\corref{cor1}}
\ead{viktor.wiberg@algoryx.com}

\author[1]{Erik Wallin}
\ead{erik.wallin@umu.se}

\author[1]{Arvid F{\"a}lldin}
\ead{arvid.falldin@umu.se}

\author[3]{Tobias Semberg}
\ead{tobias.semberg@skogforsk.se}

\author[3]{Morgan Rossander}
\ead{morgan.rossander@skogforsk.se}

\author[1,4]{Eddie Wadbro}
\ead{eddie.wadbro@kau.se}

\author[1]{Martin Servin}
\ead{martin.servin@umu.se}

\cortext[cor1]{Corresponding author}

\address[1]{Ume\aa\ University, SE-90187, Ume\aa, Sweden}
\address[2]{Algoryx Simulation AB, Kuratorvägen 2, SE-90736, Umeå, Sweden}
\address[3]{Skogforsk (the Forestry Research Institute of Sweden), SE-75183, Uppsala, Sweden}
\address[4]{Karlstad University, SE-65188, Karlstad, Sweden}


\begin{abstract}
We explore sim-to-real transfer of deep reinforcement learning controllers for a heavy vehicle with active suspensions designed for traversing rough terrain.
While related research primarily focuses on lightweight robots with electric motors and fast actuation, this study uses a forestry vehicle with a complex hydraulic driveline and slow actuation.
We simulate the vehicle using multibody dynamics and apply system identification to find an appropriate set of simulation parameters.
We then train policies in simulation using various techniques to mitigate the sim-to-real gap, including domain randomization, action delays, and a reward penalty to encourage smooth control.
In reality, the policies trained with action delays and a penalty for erratic actions perform nearly at the same level as in simulation.
In experiments on level ground, the motion trajectories closely overlap when turning to either side, as well as in a route tracking scenario.
When faced with a ramp that requires active use of the suspensions, the simulated and real motions are in close alignment.
This shows that the actuator model together with system identification yields a sufficiently accurate model of the actuators.
We observe that policies trained without the additional action penalty exhibit fast switching or bang-bang control.
These present smooth motions and high performance in simulation but transfer poorly to reality.
We find that policies make marginal use of the local height map for perception, showing no indications of predictive planning.
However, the strong transfer capabilities entail that further development concerning perception and performance can be largely confined to simulation.
\end{abstract}

\begin{keyword}
autonomous vehicles \sep rough terrain navigation \sep machine learning
\sep sim-to-real \sep reinforcement learning \sep heavy vehicles
\end{keyword}

\maketitle

\section{Introduction}
As deep reinforcement learning (DRL) evolves towards a valuable method for control in rough terrain, research has mainly focused on lightweight robots with electric motors~\cite{hu2021sim}.
While electric motors may not be well suited for tasks that involve heavy lifting or transport under harsh conditions, hydraulic systems are particularly effective.
Hydraulic actuators are durable, can deliver high forces or torques, and can withstand high impulses during operation.
The main disadvantage when it comes to controller design is their complexity, often with the presence of coupling effects, making them challenging to model~\cite{egli2022general,koivumaki2015stability}. 
The modelling challenges are enhanced when dealing with heavy vehicles in harsh environments, such as in construction, mining, and forestry, where data collection is expensive and time-consuming.
Simulators are an attractive option to train controllers as they alleviate the problem of time, safety, and cost.
But, as has been well documented, DRL controllers that work well in simulation rarely perform at the same level in the real world~\cite{zhao2020sim,dulac2021challenges,salvato2021crossing}.
This reality gap for heavy vehicles with hydraulic actuators has yet to be explored.

To make the transition towards a higher level of autonomy, heavy vehicles designed to operate in rough terrain will require advanced control methods.
In forestry, novel concepts are emerging that aim to increase vehicle mobility and act more gently on the terrain~\cite{gelin2020concept}.
A promising approach is to use active suspensions to overcome obstacles and distribute the machine's weight to minimize soil disturbance and the risk of overturning~\cite{lundback2024rubber}.
The concept of active suspensions for a full-scale forestry machine is a non-trivial design and control problem.
In previous work, we developed a controller for a wheeled forestry vehicle with active suspensions using DRL~\cite{wiberg2021control}.
The controller learnt to utilize a local height map for perception and make adequate decisions regarding if to drive straight over or circumvent obstacles.
Although the controller presented the desired characteristics on challenging terrain, including 3D reconstructions from actual forest environments, it was never tested outside the simulation environment.

In the context of rough terrain, most work on DRL controllers that attempt transfer from simulation to reality use applications from legged locomotion.
Legged locomotion is challenging due to the fast actuation, torso balance, and development of various gait patterns.
The ability to perceive the terrain and adapt the gait before ground contact is a crucial aspect of fast locomotion and robustness to terrain variations~\cite{agarwal2023legged,margolis2024rapid}.
There is a resemblance between heavy vehicles with active suspensions and legged locomotion in their need to use perception for planning and control to select a suitable path and adapt the contact organs to the terrain.
However, they differ in that hydraulic actuators rely on slow actuation, and forestry vehicles have a low centre of mass, making them relatively insensitive to the balance point.
Instead, the complication lies in modelling the hydraulic driveline.
If the driveline is modelled appropriately, there is reason to believe that the successful methods for sim-to-real transfer utilized in legged locomotion should have the same effect on wheeled vehicles with controllable suspensions and slow actuators.

A common approach to bridge the reality gap in DRL is to apply domain randomization to the simulation environment~\cite{zhao2020sim,kirk2023survey}.
The idea is to randomize parameters in the simulation that are uncertain in the real world, either because they cannot be measured or to account for unmodelled physics.
Although randomizing simulation parameters related to the dynamics, such as joint and contact friction, mass, and observation noise has been shown to produce robust controllers~\cite{peng2018sim,miki2022learning,choi2023learning}, this approach may compromise specialization in favour of generalization.

Combining domain randomization with system identification helps to balance specialization and generalization~\cite{tan2018sim}.
System identification involves determining simulation parameters that match the physical system's response to specific control signals, such as impulses or steps.
This real-to-sim approach yields a more accurate model of actuator dynamics, which is often essential for successful transfer to reality~\cite{ibarz2021train}.

Related to the model of the actuators, system latency presents additional challenges that significantly affect transfer capability.
For hydrostatic drivelines with coupling effects, the delay between the control signal and the actuator's response is large and depends on the configuration and current effort of other actuators.
Consequently, the delays are hard to predict, even through system identification.
Several studies have addressed the issue of system latencies by including the previous control in the observation space and extending the DRL controller with memory~\cite{margolis2024rapid,haarnoja2018soft,xiao2020thinking}.
Memory provides the policies with enough information to learn about when a previous action takes effect.
Recurrent neural networks, such as long short-term memory (LSTM), are an option to add memory~\cite{hu2021sim}.
Another option, which is simpler to implement yet equally effective, is to augment the observation space with several windows of past observations~\cite{ibarz2021train}.





\begin{figure}
    \centering
    \includegraphics[width=0.9\columnwidth]{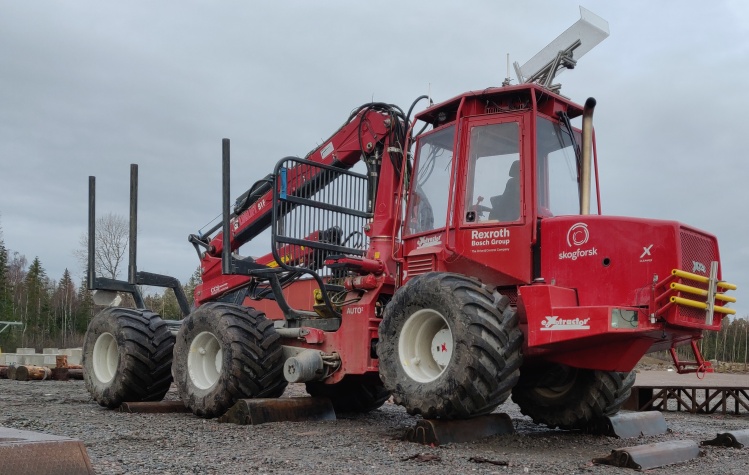}
    \caption{Xt28 forwarder on the vibration course.}
    \label{fig:xt28-vibcourse}
\end{figure}

To gain insights into the transfer capabilities of DRL controllers for heavy vehicles with hydraulic actuators, we train policies in simulation and deploy them on the physical counterpart.
The vehicle, featuring eight degrees of freedom for continuous control, is a novel concept dedicated to forestry with active suspensions and a hydrostatic driveline, see Fig.~\ref{fig:xt28-vibcourse}.
In a multibody dynamics framework, we model the actuators using kinematic constraints and identify the model parameters using system identification.
The policies undergo training on rough terrain with obstacles, where we apply domain randomization for observation noise and actuator delays.
To evaluate the transfer to reality, we deploy four distinct policies in real-world scenarios,
including simple driving scenarios, a vibration course, and ramps requiring active use of the suspensions to prevent chassis collision.
We analyze to what extent the policies use the visual input locally and for predictive planning.

\section{Real platform, Xt28}\label{sec:real-platform}
The Xt28 vehicle is a six-wheeled full-scale concept machine with a design different from typical forwarders in forestry~\cite{gelin2020concept}.
It has two articulation joints for steering instead of one and six wheels without tracks or bogies, each with an actively articulated suspension.
A diesel engine powers the drivetrain, propelling three hydraulic pumps.
Two of these pumps drive the transmission, while the third supplies power to the hydraulics for the active suspensions, steering, and crane.
Table~\ref{tab:xt28} summarises the machine's properties.
\begin{table}
    \centering
    \caption{Properties of the Xt28 forwarder prototype.}
    \label{tab:xt28}
    \begin{tabular}{l l}
        \hline
        Machine total weight, full load & \SI{31000}{kg} \\
        Load capacity             & \SI{14000}{kg} \\
        Engine maximum torque     & \SI{1300}{Nm} \\
        Engine rpm at maximum torque & \SI{1500}{rpm} \\
        Working rpm                  & \SI{1200}{rpm} \\
        Engine power at working rpm  & \SI{150}{kW} \\
        Suspension range      & \SI{0.50}{m} \\
        Wheel radius          & \SI{0.74}{m} \\
        \hline
    \end{tabular}
\end{table}




\subsection{Hydrostatic transmission}
\begin{figure}
    \centering
    \includegraphics[width=0.9\columnwidth]{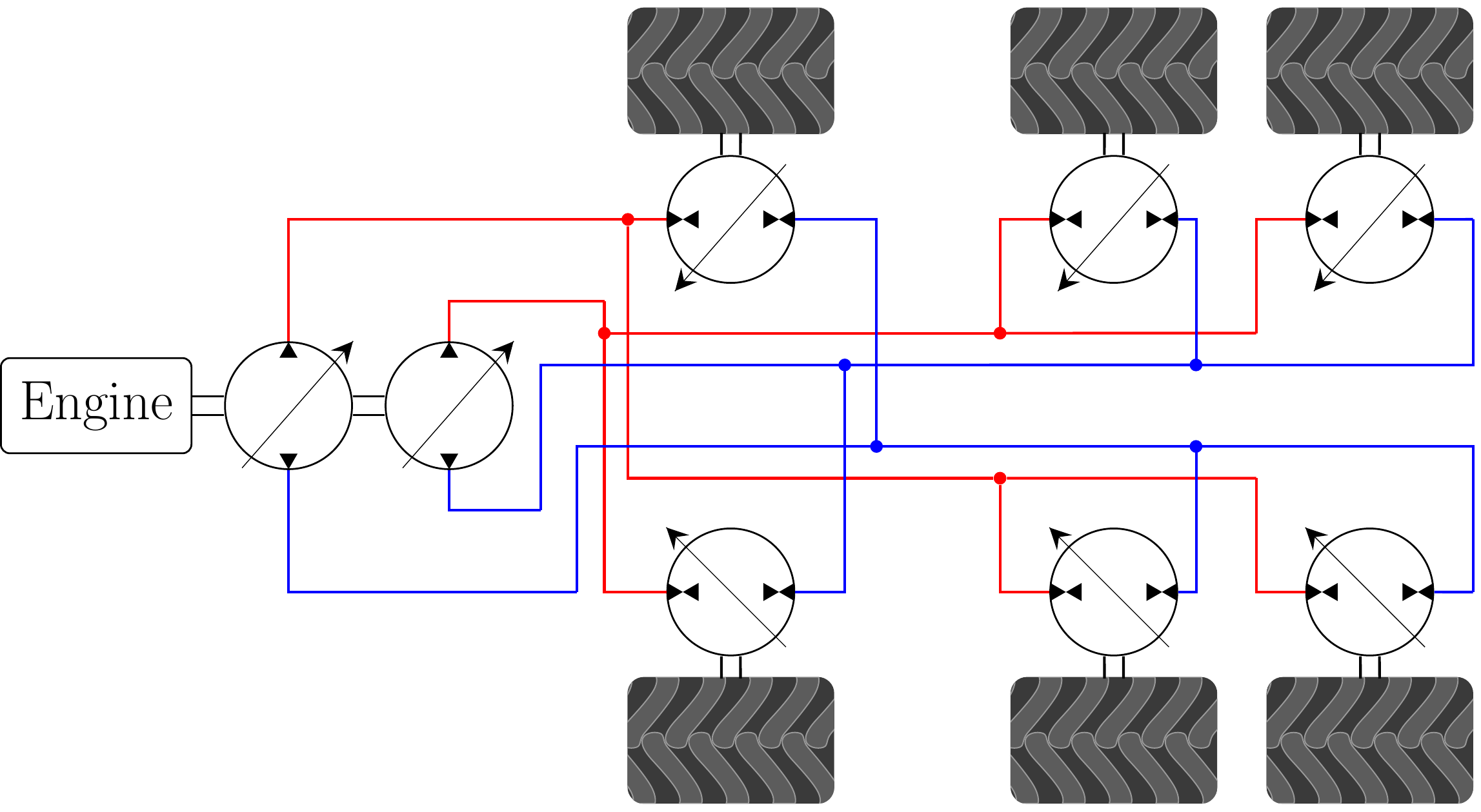}
    \caption{Circuits of the hydrostatic transmission.}
    \label{fig:xt28-transmission}
\end{figure}

Each of the two pumps in the transmission forms a closed circuit with three wheel-motors connected in parallel, see Fig.~\ref{fig:xt28-transmission}.
One wheel in the front frame and two on the opposite side in the middle and rear frame form one circuit, respectively.
The wheel-motors and the pumps all have variable displacements.
One Electronic Control Unit (ECU) controls the displacements to achieve target speed and accomplish differential behaviour during turning~\cite{dell2015modelling}.
The throttle percentage sets the target speed.
All wheels have rotary encoders, which allow measuring both speed and direction.

\subsection{Steering}
\label{sec:steering}
Push-pull pairs of hydraulic cylinders actuate the steering, with one pair in each articulation joint.
Both joints have sensors in the push-pull cylinders to allow the steering angle to be measured. 
Steering is controlled through the front joint and is regulated using a PI controller with gains $k_\mathrm{p}=0.025$, $k_\mathrm{i}=0.2~\mathrm{s}^{-1}$.
An additional PI controller ensures the angle of the rear joint is \SI{56}{\percent} of the front joint's, where the relation is determined geometrically.
We tune the rear controller and use a small delay relative to the front to achieve a trade-off between positioning accuracy and the forced pull on the machine's rear parts.
The steering also has a speed threshold that restricts the machine from turning whenever its forward speed is below $\SI{0.3}{km/h}$.

\subsection{Pendulum arm control} \label{sec:pend_arms}
The Xt28 has a dedicated ECU for controlling the pendulum arms. 
The hydraulic pump and valves that are responsible for the suspensions have a hydraulic load-sensing control system.
In load-sensing systems, the pump displacement is controlled through a hydraulic feedback line from the actuator valves~\cite{yan2022}.
Ideally, this ensures a constant flow and makes the system more adaptive to changes in load.
The speed of the cylinder is proportional to the flow, and the flow correlates to the valve control signal.

All arm cylinders are equipped with pressure sensors on both the high- and low-pressure sides.
The pressure sensors allow for a differential pressure measurement of the cylinder, which is used to estimate the ground force of each wheel. 

We have two modes for controlling the pendulum arms: conventional control, which achieves automatic active suspensions, and manual active suspensions.
A passive controller runs concurrently with either the automatic or manual active suspensions.
The mode we expose to the DRL controller is manual active suspensions.
We use the conventional controller for active suspensions during experiments to compare it with the DRL controller.

\subsubsection{Passive suspensions}
\begin{figure}
    \centering
    \includegraphics[width=0.9\columnwidth]{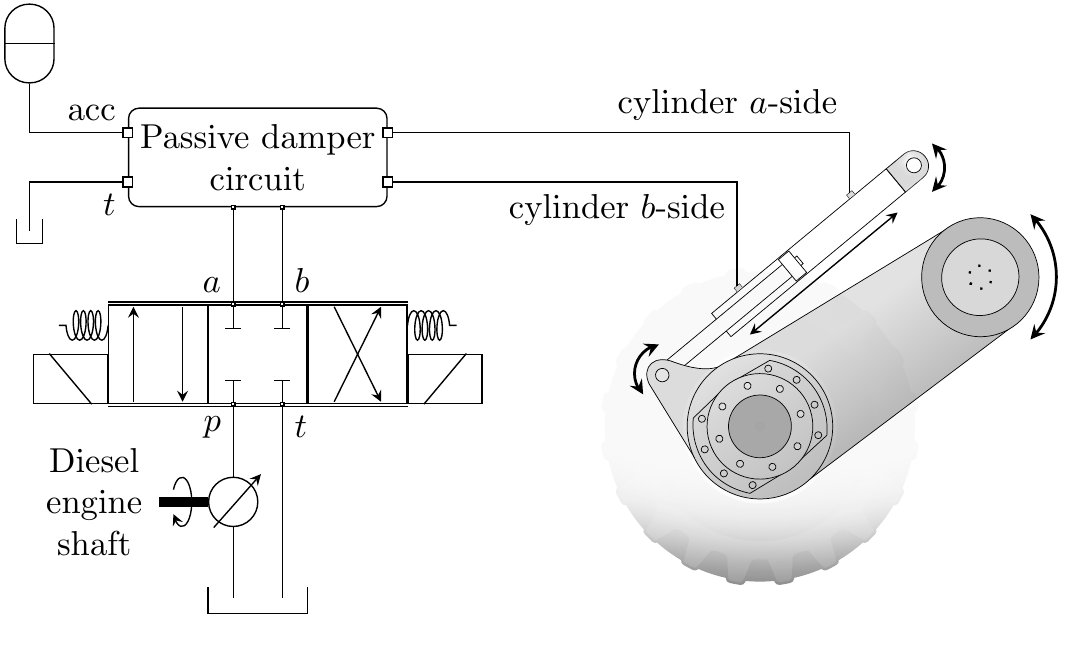}
    \caption{Simplified hydraulic circuit for the suspension of one pendulum arm. The direction control valve above the diesel engine shaft controls the fluid flow to the actuator.
    Each of the three boxes represents a different valve position, where the centre box indicates the neutral position with both ports closed.
    The pump and tank are common.}
    \label{fig:xt28-hydralic_curcuit}
\end{figure}

The passive mode is based on small pre-pressured hydraulic accumulators, one for each wheel connected to the high-pressure side of each cylinder.
The accumulators provide some slack in the system and can absorb the highest impulse forces.
Fig.~\ref{fig:xt28-hydralic_curcuit} illustrates the hydraulic circuit for a one-wheel suspension.

\subsubsection{Automatic active suspensions}
The system with automatic suspensions has a customized implementation that runs on the dedicated ECU unit for the pendulum arms.
The controller has a cascaded design in three levels:
The top level has three P-controllers for roll, pitch, and height along with a fourth controller for ground pressure.
The ground pressure controller aims to evenly distribute the normal force among all wheels and requests a force for each arm, considering the uneven weight distribution of the machine.
The pitch controller inputs a measurement of the vehicle's pitch angle and outputs a requested force change between the front and rear arms. 
Similarly, the roll controller inputs the vehicle's roll angle and outputs a requested force change between the left and right side arms.
The height controller uses the average height of the arms as input and outputs a requested force change of all arms.
The outputs of the four subsystems are expressed in terms of requests for each wheel arm, and the resulting setpoint for each arm is simply a weighted sum of all four controller requests.
The middle level of the controller is a PID controller that converts the forces into currents corresponding to valve openings.
At the lowest level is a hardware controller for electric output currents built into the ECU to ensure the setpoints for the hydraulic valve solenoids.

\subsubsection{Manual suspensions}
The machine supports ``manual suspensions'', which allow direct control of the valves of pendulum cylinders.
A single value per arm is used to move the wheel downwards, upwards, or keep still.
To provide position control of the wheel suspensions, PI controllers convert the target cylinder positions to valve openings.
The cylinder positions, that have a range $[0.0, 0.5]~\mathrm{m}$, are measured and used as feedback.
The PI controllers were tuned on the physical machine to achieve a satisfactory response without overloading the hydraulic system.
The resulting gains are $k_\mathrm{p} = 0.004~\mathrm{mm}^{-1}$, $k_\mathrm{i} = 0.0001~\mathrm{mm}^{-1}\mathrm{s}^{-1}$ with a windup limit of 100 as a maximum limit for the error integral.



\subsection{Pose and velocity measurements}
The machine has a global navigation satellite system (GNSS) with dual antennas on the cabin's roof. 
The system uses two simpleRTK2B GNSS modules from Ardusimple, each comprising ZED-F9P5 modules from u-blox.
In a moving base RTK setup, these modules connect to a local base station for correction.
The system provides positional data with the capability of cm accuracy.
It also provides velocity and heading.
An IMU is placed in the vehicle's front frame to measure the roll and pitch angles.

\subsection{Communication}
To read sensors, process data, and control the machine, a hardware interface node is implemented in Robot Operating System (ROS), release Melodic Morenia~\cite{quigley2009ros}.
The node runs on an onboard Nvidia Jetson and allows easy access to pose and velocity measurements and the vehicle joints using ROS topics.
The GNSS is connected to the Jetson with USB and exposed to ROS using the package KumarRobotics \texttt{ublox}.
ROS topics regarding joint control and sensing are communicated through the pendulum arms, steering angles, and wheels via the onboard ECUs through CAN-bus interfaces.
The ROS network is deployed onboard the machine using a 1~Gbit/s Ethernet connection.

\section{Simulation environment}
The simulation environment consists of rigid bodies interacting through frictional contacts and kinematic constraints using the physics engine AGX Dynamics~\cite{Agx2022}.

\subsection{Terrains}
To represent terrains, we use height maps on a uniform grid with 0.1~m resolution that we interpolate using triangular, piecewise planar elements, to form a continuous surface.
The geometric mesh is assigned to a static rigid body that represents the ground.
We use Perlin noise~\cite{perlin1985image} to generate terrains with natural variation and embed semi-ellipsoids to represent boulders.

The terrain-wheel contact properties determine how the vehicle interacts with the terrain.

\subsection{Vehicle model}
The vehicle model is based on a CAD drawing of the original vehicle and has 14 actuated joints.
Although it is possible to simulate hydraulic circuits and electronics explicitly, this approach would increase the number of simulation parameters to calibrate and slow down the simulation.
Instead, we model the actuator dynamics using 1D constraints.

To achieve steering, we use hinge joints with 1D motors in place of the push-pull pairs at the articulation joints of the real machine.
To simulate the hydraulic circuits illustrated in Fig.~\ref{fig:xt28-transmission}, we model the transmission as two cross-connected differentials connected to a central engine.
The engine tries to maintain a target rotational speed of the two input shafts that connect it with the differentials.
We represent each differential as a kinematic constraint that distributes torque over three output shafts, ensuring that the average speed matches that of the input shaft.
We associate the output shafts with their corresponding wheel axis of rotation.
For computational reasons concerning contact detection, we treat the wheels as rigid with spherical shapes.

Linear joints with compliant 1D lock constraints actuate the suspensions.
The lock limits the remaining degree of freedom to a fixed position in a \emph{soft} way to imitate the hydraulic accumulator's smoothing and dampening effect.
This modelling approach for the passive suspensions also aims to compensate for the simplified wheel model.
Shifting the lock's rest position along the cylinder's axis enables manual position control of the suspensions and effectively extends or contracts the shaft.

To achieve position control similar to the real vehicle, we simulate the PI controllers for suspension and articulation.
These controllers offer the same functionality as the real actuators and were tuned to make the model agree with the physical machine in step response experiments.
Therefore, the controllers also provide a means for model tuning between the simulated and real machines.





\section{System identification}
To accurately model the actuator dynamics, we design test scenarios that are run on both the real and simulated platforms.
The simulation parameters are tuned manually until a given control signal produces a response similar enough to the actual machine.
We put special effort into calibrating the model's control of the pendulum arms and the articulated steering.

In total, we tune some 50 parameters during calibration.
There are 14 parameters involved in the articulated steering, which affects both the front and rear hinges.
These parameters include PID gains and ranges for angles, torques, and angular velocities.
The pendulum arm model has 24 tunable parameters, including PID gains, actuator range, different speed limits for contraction and extension, lock constraint compliance and damping, force range limits, force thresholds, delays, and drift speed.
The remaining parameters include the engine torque range, wheel axes rolling resistance, and properties of the wheel-ground contact material, such as Young's modulus, restitution, friction, and damping coefficients.

\subsection{Pendulum arms}\label{sec:sys:arms}
The pendulum arms' actuator dynamics is characterised by interactions between the manual and passive suspension systems running in parallel.

\begin{figure}
    \centering
    \includegraphics[width=0.95\columnwidth]{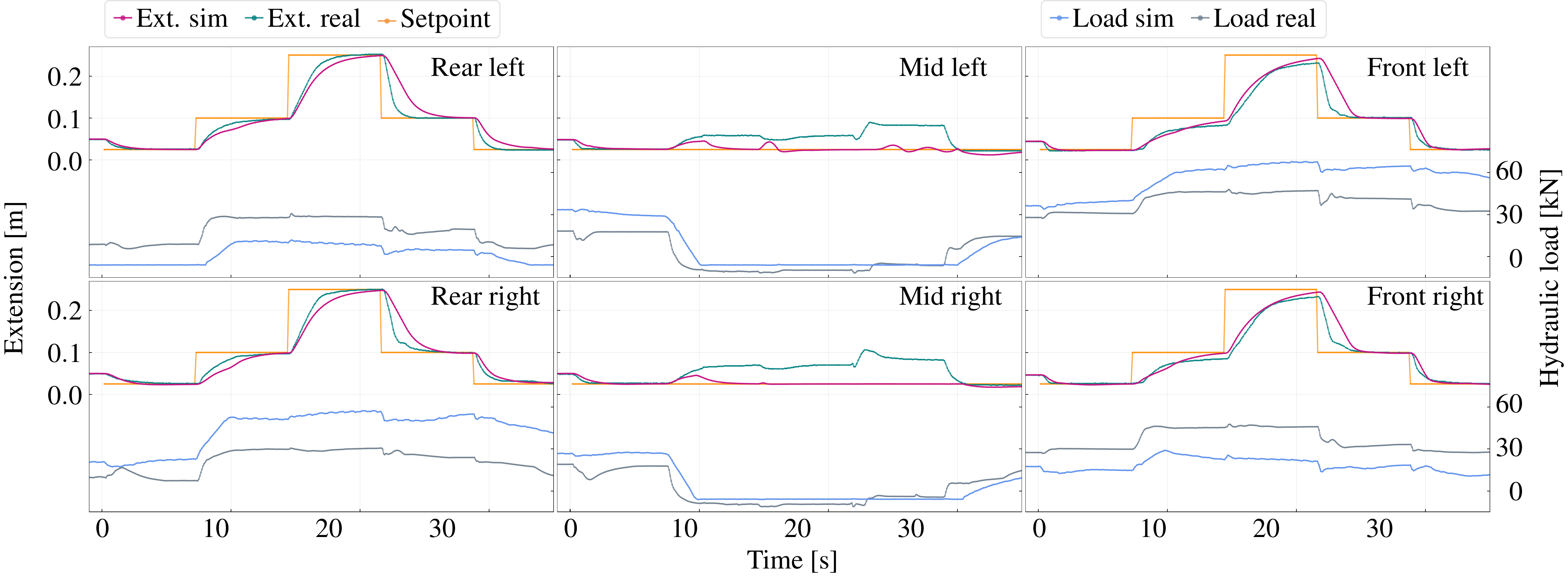}
    \caption{Step responses in arm extension and vertical load per arm in simulation compared to those of the real machine.}  
    \label{fig:cal_run_014_arms}
\end{figure}

Fig.~\ref{fig:cal_run_014_arms} shows an example of a quasi-static calibration test of the pendulum arm control.
Starting at standstill with the pendulum arms fully contracted, we extend the front and rear hydraulic cylinders while keeping the middle pair contracted.
The front and rear cylinders go from minimal to maximal extension and back again in two steps.
Note how the real machine struggles to keep the middle pair of arms in place when they are not supported by the ground.
Looking at the arm extension curves in this scenario, we see that the 
real machine is able to use the weight of the chassis to rapidly contract its arms.
One can also note that the real machine is faster to extend its rear arms than the front ones.
This is explained by the machine being front-heavy when running with an empty bunk and hydraulic coupling effects.
To overcome these issues, we considered using different PID gains for each of the three sections as well as different gains for extension and contraction.
Ultimately we decided against this to avoid adding more parameters to the model and risk overfitting.

\begin{figure}
    \centering
    \includegraphics[width=0.6\columnwidth]
        {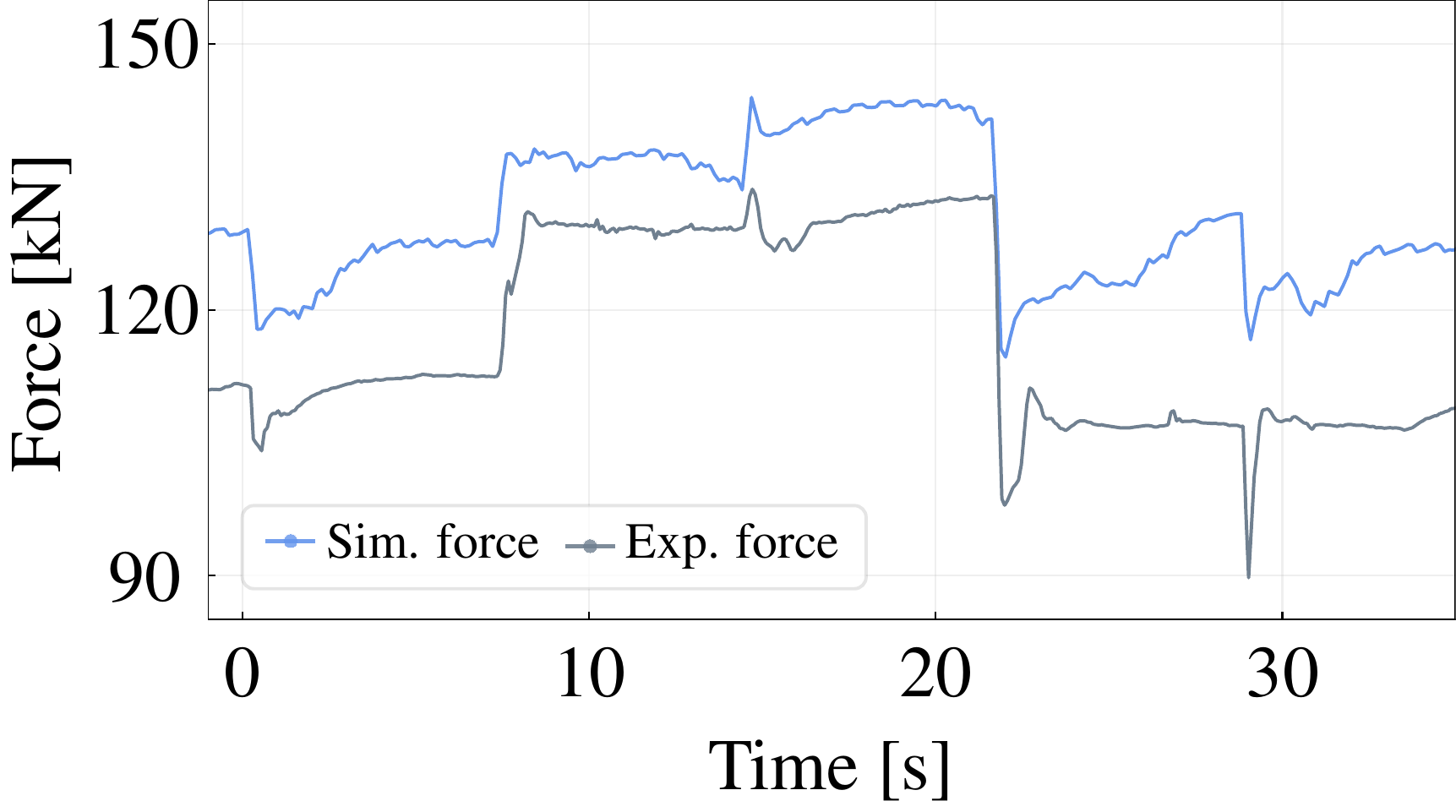}
    \caption{Total load from simulation compared with measurements from the real-world experiments. The load for each curve corresponds to the sum of vertical forces over all six wheels.
    The test scenario is the same as in Fig. \ref{fig:cal_run_014_arms}.}  
    \label{fig:cal_run_014_force_sum}
\end{figure}

Since distributing the vehicle's weight on any more than three arms is an underdetermined configuration, it is significantly more difficult to reproduce the load on each arm than to reproduce the cylinder extension.
During calibration, we typically settle for agreeing on overall trends such as the ones shown in Fig.~\ref{fig:cal_run_014_arms} and instead try to replicate the sum of the forces on all arms, as seen in Fig.~\ref{fig:cal_run_014_force_sum}.
The total vertical force on the arms is consistently larger in simulation than in reality; this is accounted for by simply rescaling the force readings before passing them to the DRL controller.

When the vehicle is at rest, we expect all vertical forces to sum up to the machine's sprung mass.
The reason behind the non-constant sums in Fig.~\ref{fig:cal_run_014_force_sum} is that the pendulum arms in the front section have mounting directions opposite to the other four, see Fig.~\ref{fig:xt28-vibcourse}.
As the front and rear arms extend, they generate a pinching motion with the ground.
When the vehicle is not in driving mode, the hydraulic fluid in the wheel motors resists rotation and increases the load on the hydraulic cylinders.
Because we rely on the hydraulic load to estimate the tyre-ground forces, we will (falsely) register this as an increase in vertical force.

\subsection{Steering \& Transmission}
\begin{figure}
    \centering
    \includegraphics[width=0.6\columnwidth]{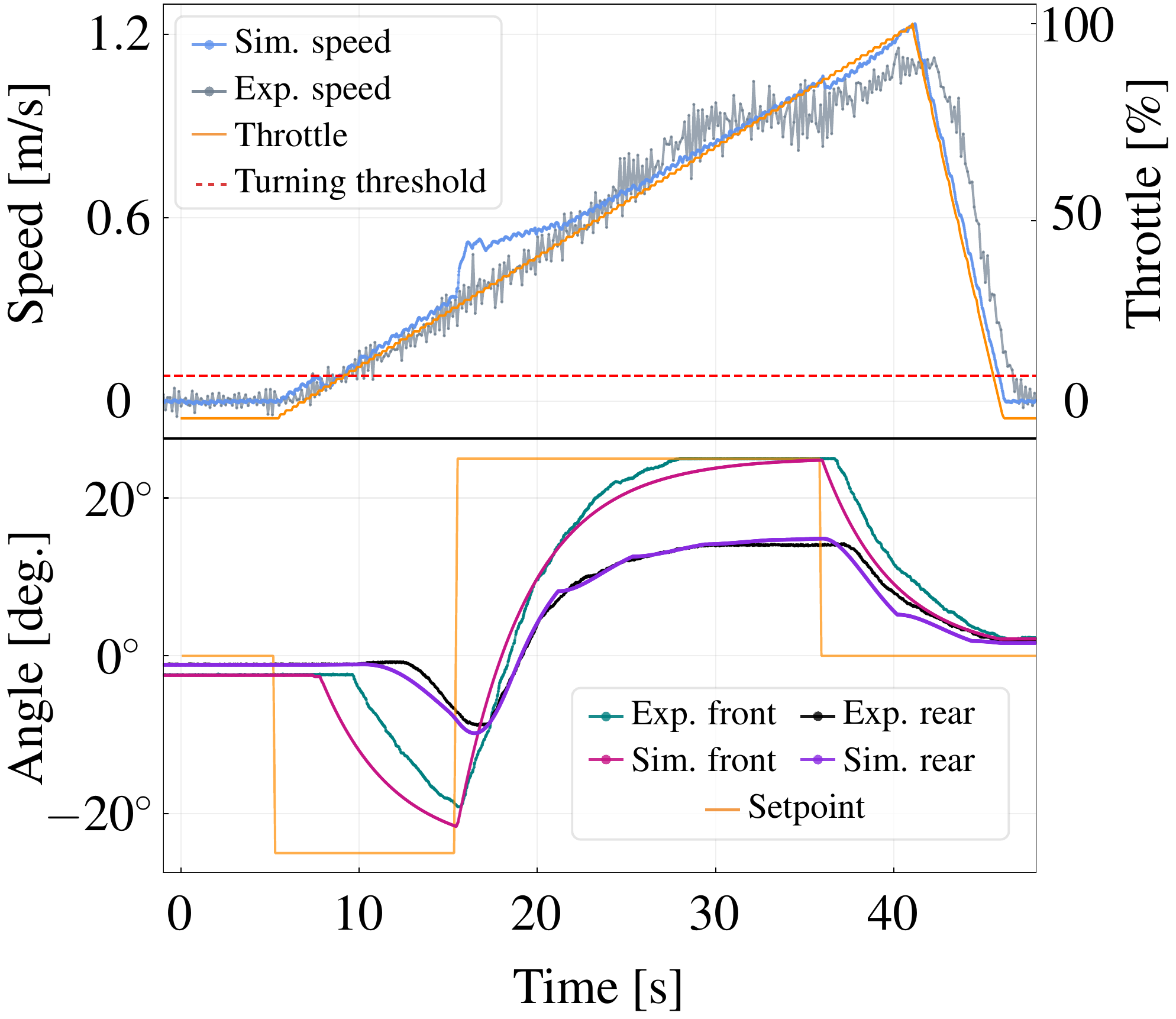}
    \caption{Comparison of articulation hinge angles and forward speed during a steering calibration test. Note that only one signal is used to control both the front and rear hinges.}
    \label{fig:cal_run_012}
\end{figure}

To model the steering, we implement the relation for the front and rear joints described in Section~\ref{sec:steering} and tune the gains in the simulated PI-controller.
Fig.~\ref{fig:cal_run_012} shows results from a test scenario with a sharp S-turn on flat terrain.
Note that the simulation does not have any actuator delays in this scenario except for the built-in delay in the rear waist hinge.
The large offset in the front angle at around $t=\SI{10}{s}$ is a result of regular delay combined with a slightly faster acceleration in simulation, making the machine reach the speed threshold for turning a few moments earlier.
The offsets at the start of the final turnaround at $t=\SI{36}{s}$ are examples of pure actuator delays.



\subsection{Model validation}
\begin{figure}
    \centering
    \includegraphics[width=0.9\columnwidth]{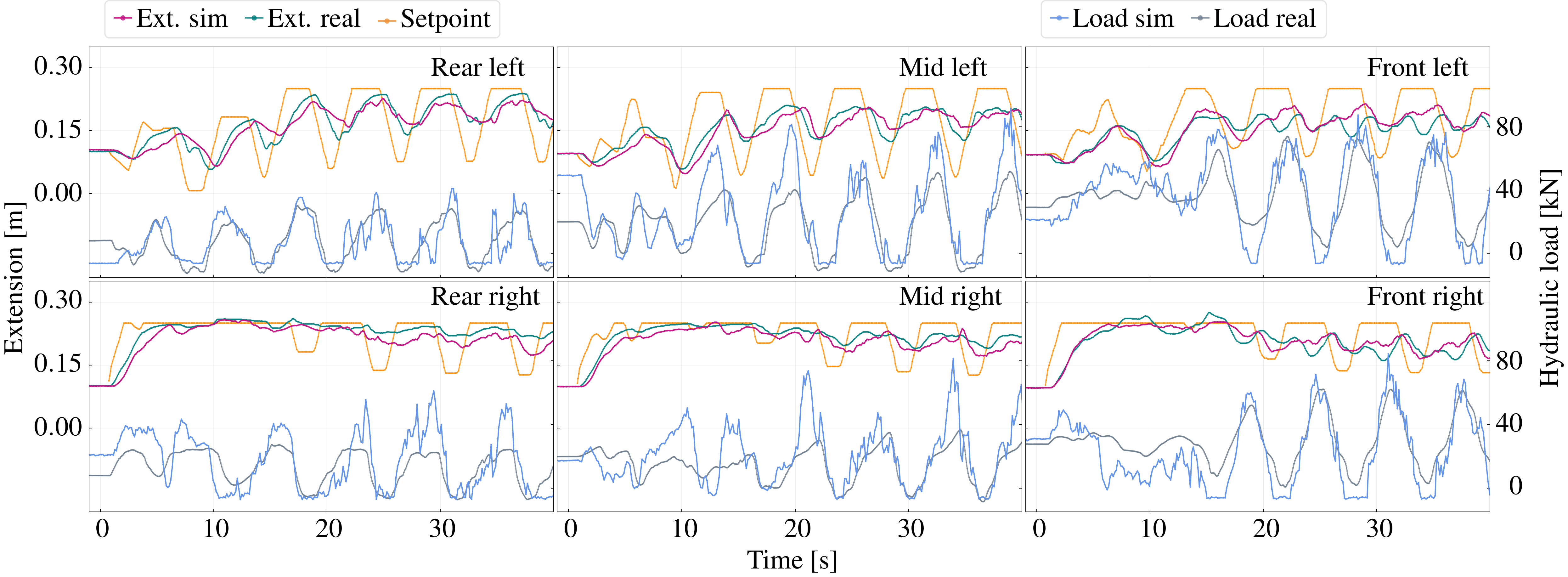}
    \caption{Comparison of arm extension and vertical load per arm sim-vs-real during an agent-controlled test run.}
    \label{fig:cal_run_037_arms}
\end{figure}

To prevent overfitting, we set aside data from a few of the calibration tests and DRL-controlled test runs and used them as a validation set.
Fig.~\ref{fig:cal_run_037_arms} shows the pendulum arm control signals and responses during a live run as the result of a prototype DRL controller.
The vehicle is commanded to drive 25~m straight ahead on flat terrain.
Despite the erratic bang-bang control signals, the responses in the simulation show good agreement with the ones from the real experiment.

\subsection{Actuator latency and sensor noise} \label{sec:actuator-latency-and-noise}
As the calibration experiments show, the vehicle's hydraulic driveline inherently leads to delays for the actuators at different timescales.
Instead of integrating these in the vehicle model, we include them when training DRL controllers.

To determine typical delays, we conduct empirical measurements by sending control signals and measuring the response time of the actuator~\cite{tan2018sim}.
Our tests reveal that the throttle delay is in the range $[0.2, 0.7]$~s, the frame articulation delay in $[0.2, 0.5]$~s, and the suspensions' delay in $[0.1, 0.2]$~s.

We assume that sensor noise follows a Gaussian distribution with a zero mean and estimate standard deviations from either data sheets or calibration experiments.
We find a standard deviation of 0.005~m for the positional data, 0.075~m/s for planar velocity, 0.025~m/s vertically, and 0.005~rad for the heading.
Terrain height maps already contain noise from the original laser scans.
From maps collected on a flat parking lot, we estimate a suitable standard deviation of 4.5~cm.
We do not model the noise from the force measurements in the suspensions' hydraulic cylinders since these are already noisy in simulation, see Fig.~\ref{fig:cal_run_037_arms}.
We also ignore noise in the measurements of suspensions' positions and the roll and pitch from the IMU.

\section{Controller}
\begin{figure}
    \centering
    \includegraphics[width=0.9\columnwidth]{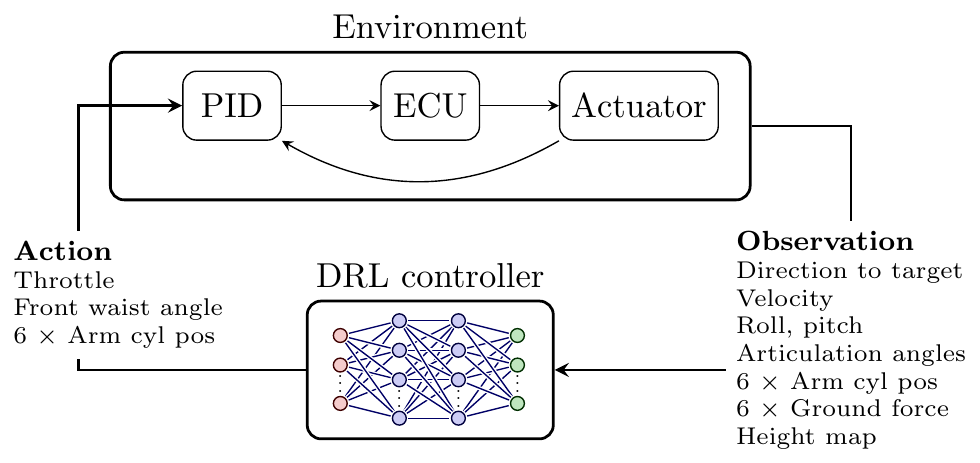}
    \caption{Overview of the control system.}
    \label{fig:control_system}
\end{figure}

The goal of the DRL controller is to safely and efficiently reach a target pose specified by a planar position and heading.
The control system is based on several components, as illustrated in Fig.~\ref{fig:control_system}.
The reinforcement learning trains the controller, which is supported by PID controllers for the different actuators.

As a safety precaution, to assure that the vehicle cannot overturn, the DRL controller is set to use at most half of the suspension range, i.e. $[0.00, 0.25]$~m.
For the same reason, we also limit the signal to the suspensions to \SI{30}{\percent} of full arm movement speed.

\subsection{Control policy}
The policy is a multivariate Gaussian represented as a neural network with parameters $\theta$ that maps state $s$ to mean action $\mu_\theta(s)$.
During exploration, the probability of selecting action $a$ in state $s$ is given by $\pi_\theta(a|s) = \mathcal{N}(a|\mu_\theta(s), \mathrm{diag}(\sigma^2))$.
The variance vector $\sigma^2$ is treated as a standalone parameter, independent of the state.

\subsection{Observation and action}
In its local frame of reference, the controller receives directions to the target pose as a planar position with heading $(x, y, \psi)$ along with velocity $v$ and the vehicle's roll $\phi$ and pitch $\alpha$
The vehicle configuration is input in terms of the piston displacement for each suspension and both frame articulation angles.
To learn about the wheel-terrain interaction, we provide the forces in the hydraulic cylinders. We scale each force input by the sum of forces during static equilibrium and clip it to $[-1, 1]$.
We also include the previous action in the observation space.

The exteroceptive input consists of a local height map that follows the vehicle position and heading in a global frame.
The heights are extracted through localization in a global map~\cite{wiberg2021control,li2020localization}.
A global map can be obtained in advance from aerial laser scans and processed offline to only include the terrain.
The local map extends $15 \times 10$~m$^2$ with a grid of $30 \times 20$ points.
Heights are taken relative to the reference frame and scaled to be in $[0, 1]$.

Given an observation, the controller outputs an 8D action in $[-1, 1]$.
The action specifies throttle, target angle for the front frame articulation, and target position for each suspension joint.

\subsection{Implementation on the Xt28}
The DRL controller is implemented in Python 3.8 and runs on a laptop placed inside the cabin.
We employ a \texttt{Rosbridge}~\cite{crick2017rosbridge} for seamless communication between the DRL controller and ROS, regardless of the ROS system, Python version, or packages involved.

To construct the observation space as the policies expect, we must deal with sensors that run on different frequencies.
All sensors run at 100~Hz except the GNSS, which has a frequency of 8~Hz and thus defines the frequency of the DRL controller.
While waiting for the GNSS, we store data from the other sensors in running lists.
When the GNSS publishes, we average over all sources with more than one value to reduce noise.
We transform the GNSS data from geodetic to local frame, which is also used to extract the local height map from the elevation model of the environment.
Finally, we map observations to the ranges used in the simulation before feeding them to the policy.
The policy outputs a vector of actions, which are then mapped to their corresponding actuators and published as ROS topics for actuation.

\section{Learning control}
To learn control policies, we use the \texttt{stable-baselines3}~\cite{stable-baselines3} implementation of PPO~\cite{schulman2017proximal}.
The architecture of the policy and value network as well as the hyperparameters are the same as in our previous study~\cite{wiberg2021control}.
The exception is the discount factor $\gamma = 0.995$ to account for longer episodes and the addition of a limit for the Kullback-Leibler divergence of 0.1 between policy updates.
We let the simulation run at 60~Hz and use a control frequency of $f_\mathrm{control} = 10$~Hz.

\subsection{Reward}
The complete reward consists of individual terms, as developed and described in detail in~\cite{wiberg2021control}, and takes the form
\begin{equation}\label{eq:reward}
    r = r_\mathrm{tar} +
        r_\mathrm{prog} r_\mathrm{speed} r_\mathrm{head} r_\mathrm{forces}
        r_\mathrm{roll}.
\end{equation}

The target reward is defined as $r_\mathrm{tar} = k_\mathrm{tar} \mathds{1}(\psi, d_t)$, where $k_\mathrm{tar}$ is a constant set to \SI{5}{\percent} of the maximum, undiscounted, episodic return and the indicator function $\mathds{1}$ evaluates to 1 at the target and 0 otherwise.
Success is defined as being within 0.3~m of the target with less than 9$^\circ$ relative heading and 7.5$^\circ$ roll angle.
To densify the signal, we reward progress toward the target as
$r_\mathrm{prog} = (d_{t - 1} - d_t)f_\mathrm{control}$,
where $d_t$, $d_{t-1}$ is the current and previous distance from the vehicle to the target projected to the horizontal plane.
The term
$r_\mathrm{speed} = \min(1, \exp[k_\mathrm{speed} (v_\mathrm{lim} - |v|)])$
encourages limited vehicle speeds, where $v_\mathrm{lim} = 0.8$~m/s, and $k_\mathrm{speed}=2$~s/m is a constant manually tuned to control the rate of reward decay for speeds above $v_\mathrm{lim}$.
Heading alignment is increasingly important as the vehicle approaches the target as
$r_\mathrm{head} = 
    \exp [-\frac{1}{2}
            \left( \psi / (d_t / k_\mathrm{d}) \right)^2 
         ]$,
where the constant $k_\mathrm{d}=5$~m is tuned in accord with the turning radius of the vehicle.
To limit ground pressure, we consider the standard deviation of normalized ground forces, $\sigma_\mathrm{forces}$.
We promote even weight distribution through
$r_\mathrm{forces} =
    \exp [-\frac{1}{2}
    \left( \sigma_\mathrm{forces} / k_\mathrm{forces} \right)^2]$,
where $k_\mathrm{forces} = 0.1$.
To avoid the risk of overturning, we define the roll reward as
$r_\mathrm{roll} = 
    \exp [-\frac{1}{2}
    \left( \phi / k_\mathrm{\phi} \right)^2 ]$,
where $k_\mathrm{\phi} = \pi / 16$.

The reward~(\ref{eq:reward}) is sufficient to train well-behaved policies but may yield a control signal that exhibits fast switching between or near the control limits.
While extremal switching or bang-bang control may be optimal in many continuous control problems~\cite{seyde2021bang}, it is unsuitable for several reasons.
Pure or near bang-bang behaviour may cause wear and tear on the equipment, unnecessary energy consumption, and excite higher-order dynamics that aggravates sim-to-real transfer.
Several works address the problem of bang-bang control from DRL controllers by introducing action regularization in the objective function~\cite{mysore2021regularizing} or by constraining the optimization problem~\cite{bohez2019value}.
However, with the risk of limiting exploration, we take the common approach of adding a penalty to the reward function.

Striving towards a smooth control signal, we use an additional term in the reward
\begin{equation}\label{eq:reward_delta}
    r' = r + r_{\Delta_a},
\end{equation}
where $r_{\Delta_a} = \Delta_a^\mathrm{T} M \Delta_a$ penalizes the difference between the current and previous action, $\Delta_a$.
The diagonal matrix $M$ holds a weight factor for each actuator.
We use $-0.01$ for the throttle and the steering, and $-0.05$ for the suspensions.

\subsection{Training} \label{sec:training}
The training uses ten simulation environments that run in parallel with different terrains.
Before an episode starts, the environment deploys the vehicle with a random heading at a random position within the unit square of the terrain origin.
We place the target 25~m away along a circular arc, sampling the relative heading uniformly within a predefined range $\pm \psi_\mathrm{range}$.
Instead of always setting the initial target heading in the radial direction of the arc, we introduce a shift uniformly sampled from $\pm \psi_\mathrm{range}/2$.

An episode has a horizon of 45~s, corresponding to 450 control steps.
We terminate the episode if the vehicle reaches the target pose or encounters a terminal condition.
A terminal condition occurs if the vehicle exceeds the roll limit of 30$^\circ$, misses the target, or has contact between anything other than the wheels and the ground.

To modulate difficulty level during training, we use a two-lesson curriculum.
In the first lesson, the focus is on basic driving skills on uneven terrains without obstacles.
This lesson facilitates sharp turns according to the target initialization with heading parameter $\psi_\mathrm{range} = \pi / 6$.
In the second lesson, we focus on using the suspensions and put less effort into turning with $\psi_\mathrm{range} = \pi / 8$.
This lesson introduces terrains with more variations and embedded obstacles to represent passable boulders.
We change the learning rate from $2.5\times10^{-4}$ in the first lesson to $1.0\times 10^{-5}$ in the second.

We train policies through the two-lesson curriculum in three ways to gain insight into how different approaches affect the reality gap.
\begin{itemize}
    \item[$\mathcal{T}_\mathrm{A}$:] Under ideal conditions using reward~\eqref{eq:reward}.
    \item[$\mathcal{T}_\mathrm{B}$:] Under ideal conditions using reward~\eqref{eq:reward_delta}, whose additional purpose is to mitigate bang-bang control.
    \item[$\mathcal{T}_\mathrm{C}$:] Including observation noise and action delays using reward~\eqref{eq:reward_delta}.
\end{itemize}

\subsubsection{Including action delays and observation noise}

We include action delays into training $\mathcal{T}_\mathrm{C}$ at the start of each episode by sampling delays for the throttle, frame articulation, and suspensions.
To sample from suitable ranges for each actuator, we rely on the time scales determined during system identification, see Section~\ref{sec:actuator-latency-and-noise}.
Depending on the sampled delay, we withhold actions several steps before passing them to the actuators, which then respond immediately.

For the controller to learn about delays, we augment the observation space with a \emph{window} of previous observations, excluding the height field data.
We use a window size of eight considering the control frequency of 10~Hz so that the largest possible delay of 0.7~s can be part of the observation history.


During training $\mathcal{T}_\mathrm{C}$, we also add independent Gaussian noise for observations of positional data, velocity, heading, and local height map readings. The noise has zero mean and standard deviations presented in Section~\ref{sec:actuator-latency-and-noise}.

\section{Results and discussion}
\begin{figure}
    \centering
    \includegraphics[width=0.8\columnwidth]{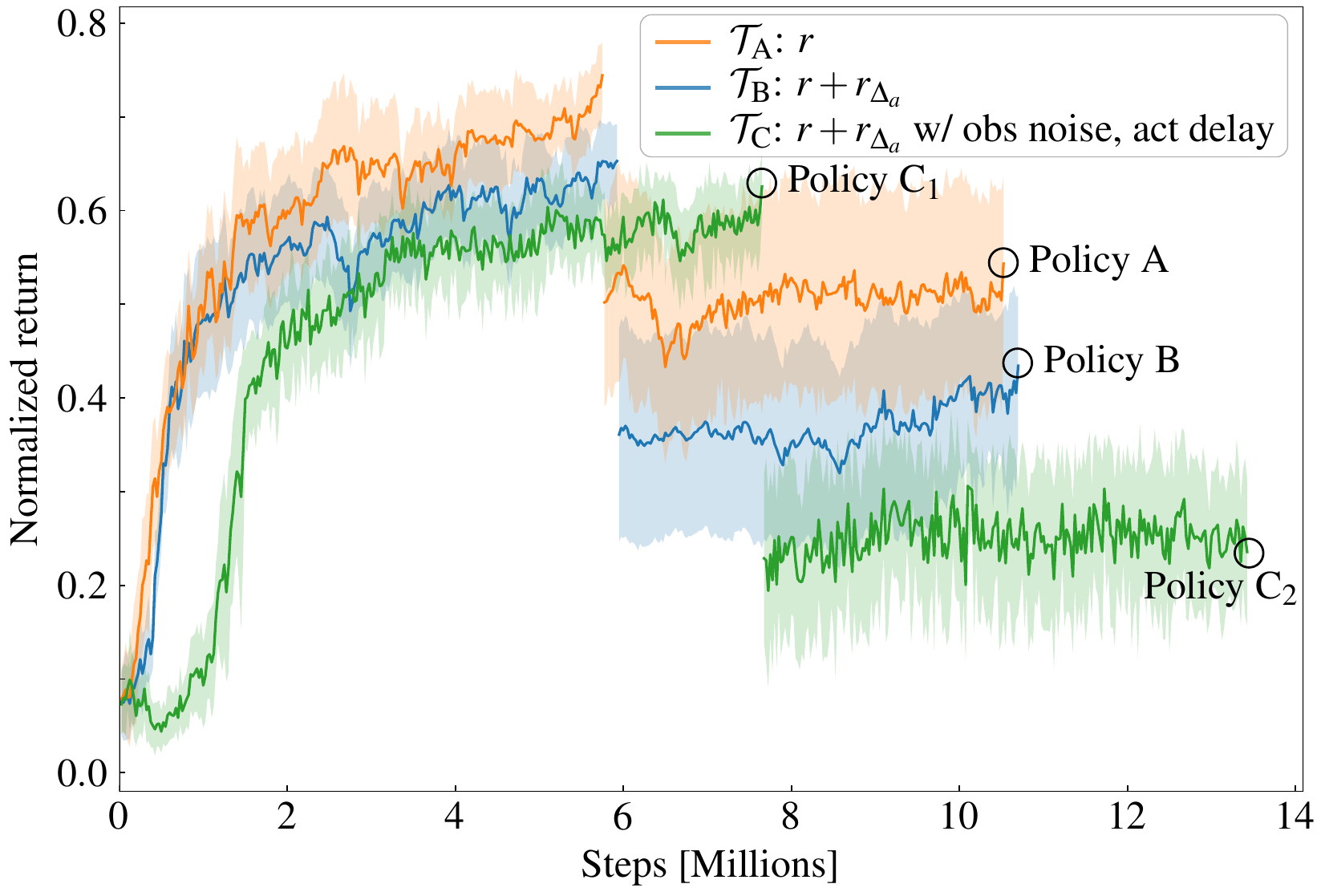}
    \caption{Learning curves plotted over the two-lesson curriculum, evaluated over 20 episodes once every \SI{25}{k} steps. The three cases $\mathcal{T}_{(\cdot)}$ correspond to training with different rewards under ideal or non-ideal conditions, see Section~\ref{sec:training}. After the first lesson, the policy from each curve with the highest evaluation return is used as a starting point for the second, indicated by a drop in return. We decided on four policies for transfer to the actual vehicle marked by A, B, $\mathrm{C}_1$, $\mathrm{C}_2$. We choose Policy $\mathrm{C}_2$ for its high success rate of \SI{95}{\percent} (not presented in the figure).}
    \label{fig:learning-curve}
\end{figure}

Training following the curriculum for $\mathcal{T}_\mathrm{A}$, $\mathcal{T}_\mathrm{B}$, and $\mathcal{T}_\mathrm{C}$ resulted in the learning curves in Fig.~\ref{fig:learning-curve}.
In general, progress during the second lesson is marginal.
The reason is either that the terrains with obstacles are too difficult or that the low ground clearance developed during the first lesson is hard to unlearn.
Without a higher ground clearance, the chassis frequently collides with obstacles that cause episode termination.

After training, we analyze the behaviour of several policy candidates in various simulated driving scenarios.
From the analysis, we decided on four policies for transfer to the actual vehicle, denoted and marked in Fig.~\ref{fig:learning-curve} as A, B, $\mathrm{C}_1$, and $\mathrm{C}_2$.
We anticipate policies $\mathrm{C}_1$ and $\mathrm{C}_2$ to demonstrate the highest transfer capabilities to the actual vehicle since they underwent training with observation noise, action delays, and reward~(\ref{eq:reward_delta}) incorporating a penalty for changes in the control signal.
We discovered that Policy~$\mathrm{C}_2$ was better than $\mathrm{C}_1$ at overcoming obstacles but degraded in terms of a smooth control signal, leading us to proceed by evaluating both of them on the physical machine.
We keep policies A and B for comparison.

The field experiments took place at Troëdsson Teleoperation Lab, run by Skogforsk and located in Jälla, Sweden.
The area consists of a large gravel parking surrounded by blocky forest terrain.
Two years before the field experiment, the entire area was scanned using airborne photogrammetry that resulted in a digital surface model with a resolution of~\SI{0.25}{m}. 
Combining the geotagged surface model with the onboard RTK GNSS allows us to extract local height maps that are part of the controller observation space.

\subsection{Basic driving}\label{sec:results:basic}
\begin{figure}
    \centering
    \includegraphics[width=0.9\columnwidth]{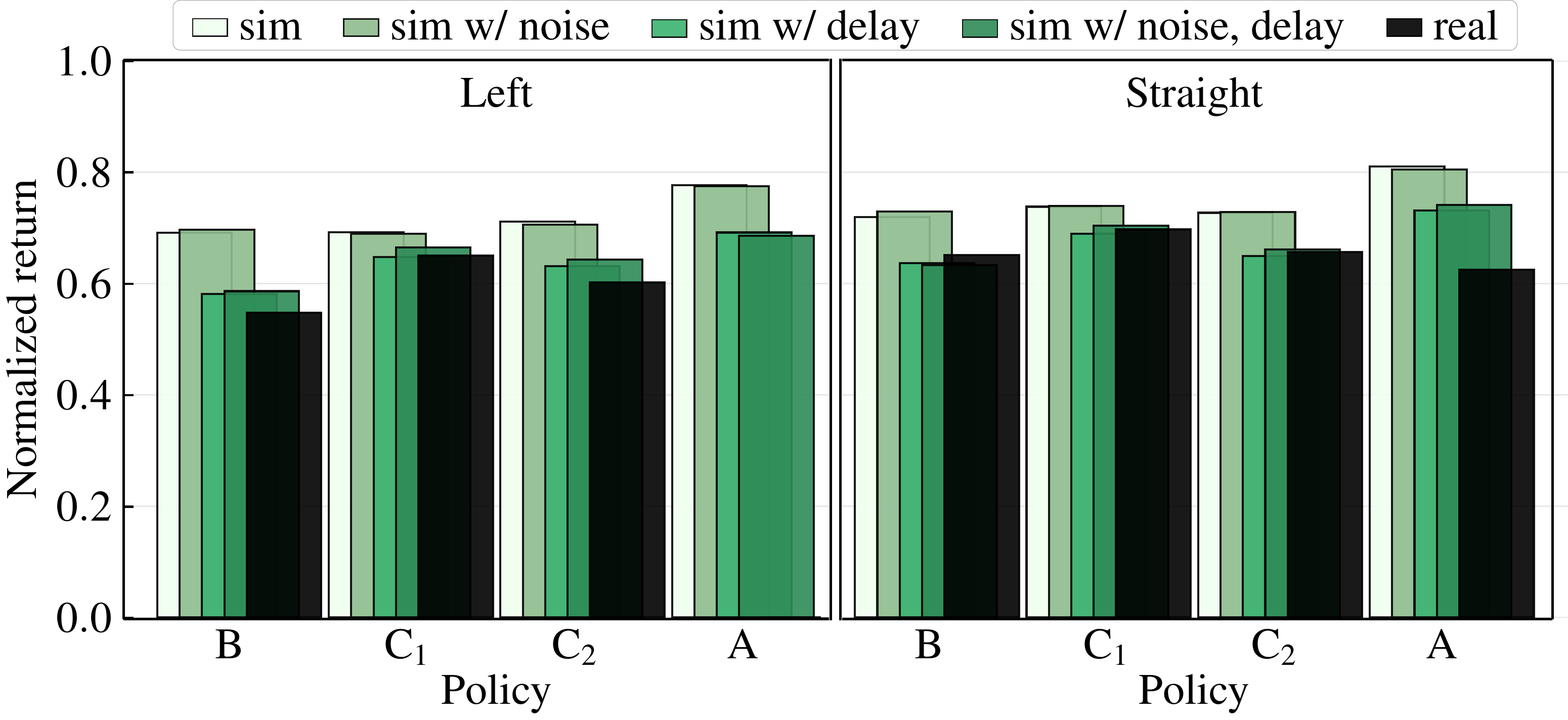}
    \caption{Policy performance in simulation compared to reality according to the base reward. The target is placed on flat ground 25~m away straight ahead or to the left, corresponding to an initial relative heading of $23^\circ$. The right case was omitted due to symmetry. Unfortunately, an error in the GNSS caused unusable data for Policy~A for left and right turns.}
    \label{fig:basic-driving}
\end{figure}

Since reward~\eqref{eq:reward} is shaped to capture the desired driving behaviour, we use it to evaluate policy performance for three scenarios on flat ground: driving straight, turning left, and right, see Fig.~\ref{fig:basic-driving}.
The simulation cases vary in complexity from perfect observations and immediate action to noisy observations and delayed actions. The real data was collected on the flat parking space.







Among the four policies, $\mathrm{C}_1$ shows the best transfer capabilities over all scenarios, while policies B and ~$\mathrm{C}_2$ are comparable.
On flat terrain, we expect Policy~$\mathrm{C}_1$ to outperform Policy~$\mathrm{C}_2$ because the latter proceeds to train on the second lesson in the curriculum, featuring rougher terrain with boulders.
This lesson causes Policy~$\mathrm{C}_2$ to trade specializing in flat terrain for handling obstacle avoidance and practising more active use of the suspensions.
Despite training under ideal conditions, the base Policy~A performs the best in simulation even with observation noise and action delays but sees a large drop when deployed in reality.
Overall, we note that adding delays to simulations has the most pronounced effect on policies that lack experience with such, while the impact of noise is minor for all policies.

Evaluating performance in terms of the reward does not justify the true experience of witnessing the field tests.
For some runs, the real machine displayed shakiness, pulsating sound from the hydraulic motors, and other unwanted behaviour that we had not anticipated and therefore not included in any reward function.

\begin{figure}
    \centering
    \includegraphics[width=0.48\columnwidth]{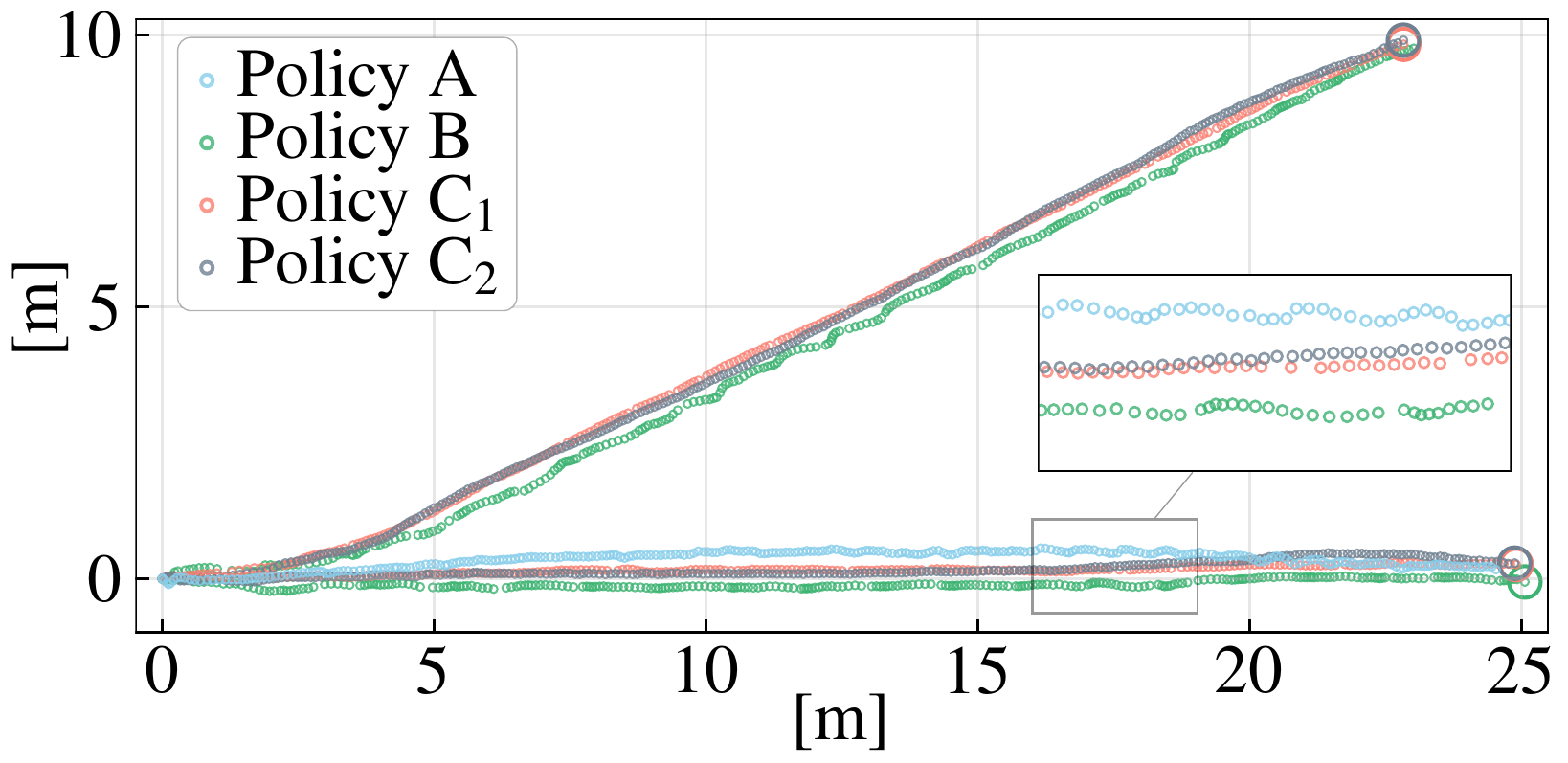}
    \includegraphics[width=0.48\columnwidth]{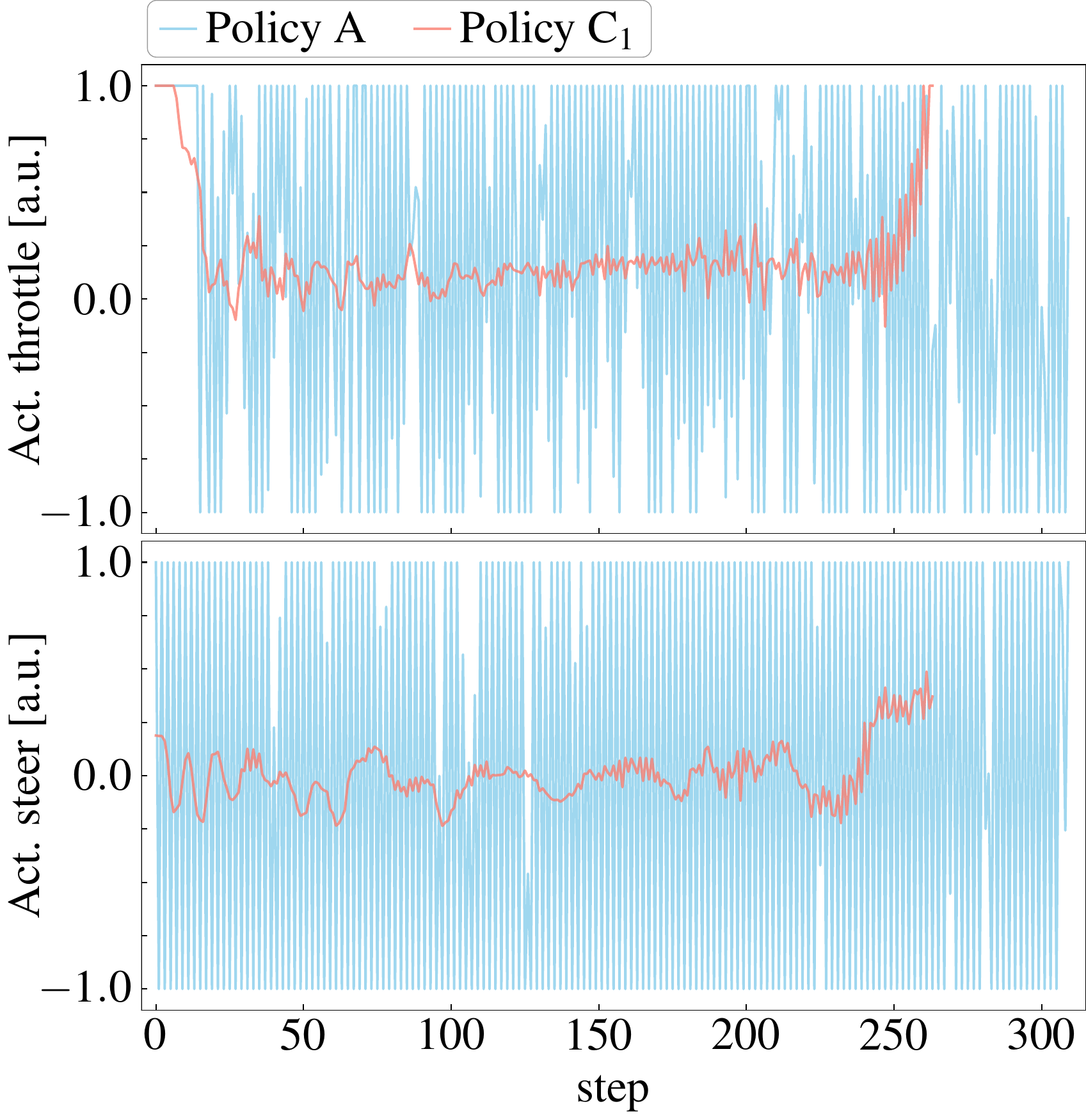}
    \caption{\textbf{Left:} Physical trajectories on even terrain, where a large circle at the endpoint indicates reaching the target. The target pose is placed 25~m away with initial relative headings $0^\circ$ and $23^\circ$, respectively. Policy~A only has one trajectory due to an error in the GNSS for the other two. \textbf{Right:} Policy action for throttle and front steering angle in the case of driving straight.}
    \label{fig:lsr-traj}
\end{figure}

In the simulations, all four policies demonstrate smooth motions, a necessary trait for real-world applications.
To study the actual motions, we plot the trajectories relative to the initial pose, see the left panel in Fig.~\ref{fig:lsr-traj}.
Policies A and B yield wiggly trajectories compared to $\mathrm{C}_1$ and $\mathrm{C}_2$ that progress towards the target smoothly.
In all three cases, Policy~$\mathrm{C_2}$ reaches the target, Policy~$\mathrm{C_1}$ achieves it twice out of three times, and Policy~B only reaches it once.
The right panel in Fig.~\ref{fig:lsr-traj} shows the control action for the throttle and steering angle with the target straight ahead.
As is clear, Policy~A exhibits bang-bang control, which works well in simulation but transfers poorly to reality.
During the experiments, we could hear the hydraulic motors struggling rhythmically to build up and release pressure.
The signal from Policy~$\mathrm{C}_1$ resulted in smooth motions and overall best behaviour from a visual and auditorial standpoint among all four policies.
The results confirm the importance of including actuator delays during training.
The results also show that policies are likely to converge to an unwanted bang-bang controller without some regularisation for actions.

\begin{figure}
    \centering
    \includegraphics[width=0.7\columnwidth]{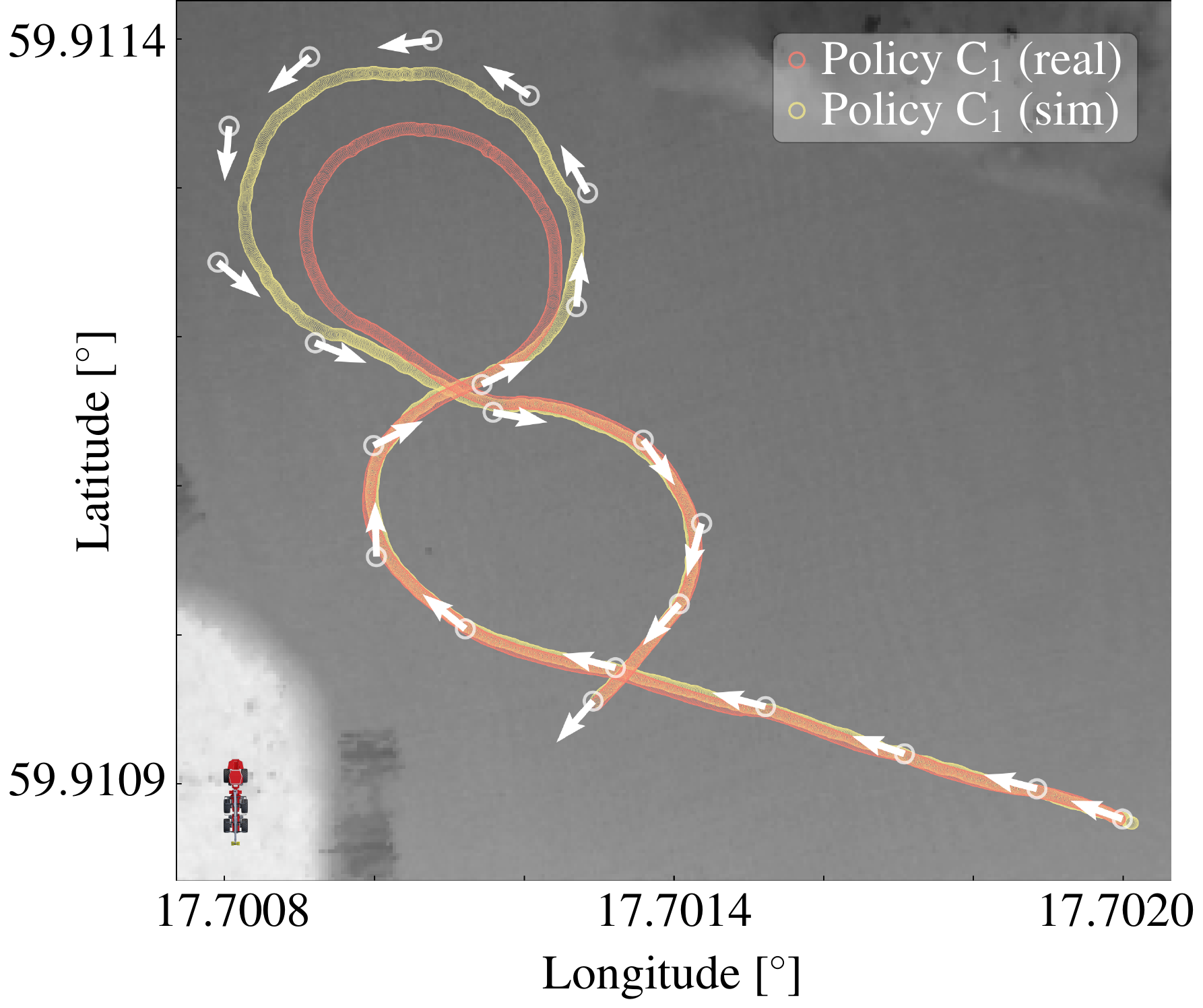}
    \caption{Real and simulated trajectories starting in the bottom right executed by Policy~$\mathrm{C}_1$ when set to follow a sequence of target poses. Target positions are drawn as white circles, and the corresponding arrows show the target heading. The vehicle is true to scale.}
    \label{fig:drive-eight}
\end{figure}

To mimic a practical use case, we test Policy~$\mathrm{C}_1$ on a sequence of targets that trace out a predefined route.
Given the initial vehicle pose, we use a simple GUI application to manually choose subsequent target poses on a 2D map of the digital elevation model.
To see how the controller handles situations incompatible with the kinematic constraints of the vehicle and unseen states, some subsequent targets require impossibly sharp turns.
The experiments are carried out in reality as well as in simulation, see Fig.~\ref{fig:drive-eight}.
In general, the controller succeeds to follow the route and trace out a smooth trajectory in both reality and simulation.
The decision-making of the policy is not straightforward to explain.
For some reason, the real case starts to deviate from the simulation at the first left turn in Fig.~\ref{fig:drive-eight}, causing the controller to end up in a challenging situation.
From that moment on, the following targets become increasingly difficult, to the point where the targets in the upper left part are missed almost immediately.
As a consequence, the controller then focuses on the next target, which exacerbates the offset compared to the simulation case.
Eventually, the controller recovers to complete the route and does so without presenting any high-risk manoeuvres.
Apart from the aforementioned deviations, the real and simulated motions overlap, showing adequate transfer capabilities in terms of driving and steering.

\subsection{Vibration course}
\begin{figure}
    \centering
    \includegraphics[width=0.8\columnwidth]{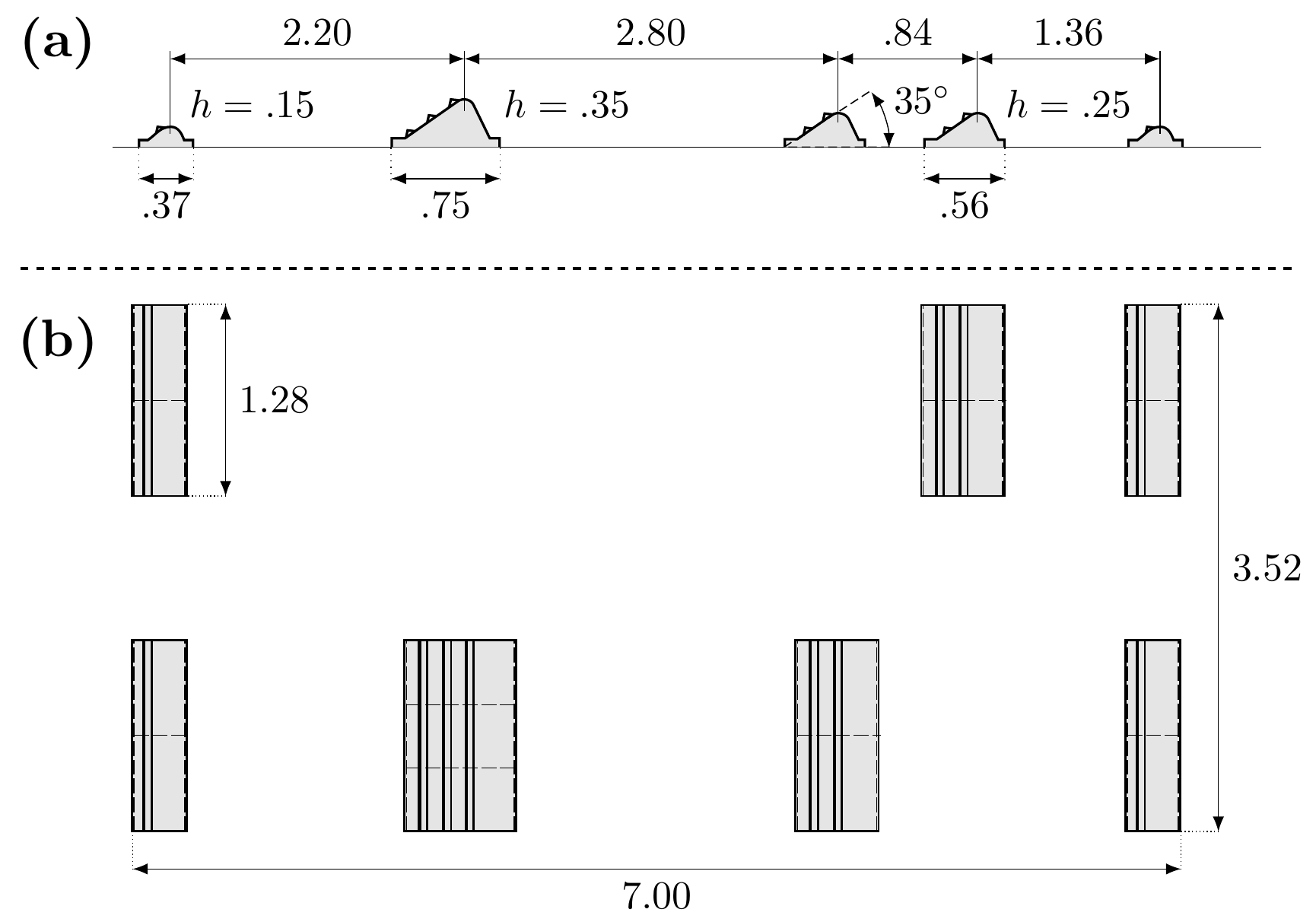}
    \caption{Illustration of the vibration course from a side view (a) and top-down view (b). The machine drives from left to right in the figure. All values are in metres.}
    \label{fig:vibration_course_schematic}
\end{figure}

To study the use of the height map information for controlling the arms, we use a \emph{vibration course}. 
It consists of obstacles, normally bolted to a concrete foundation, following a standardized design and layout to compare the performance of different forwarders, see Fig.~\ref{fig:vibration_course_schematic}.
We use approximately the same layout and attach the obstacles to the gravel ground.

During the field experiments, we compare two cases of driving through the vibration course, with and without adding obstacles to the digital surface model.
In the latter case, the controller only sees the ground and must rely fully on the proprioceptive information to physically feel the obstacles.
There was no apparent difference between the two runs, although noisy measurements with several obstacles and variations in initial conditions make the results difficult to analyze.

\begin{figure}
    \centering
    \includegraphics[width=\columnwidth]{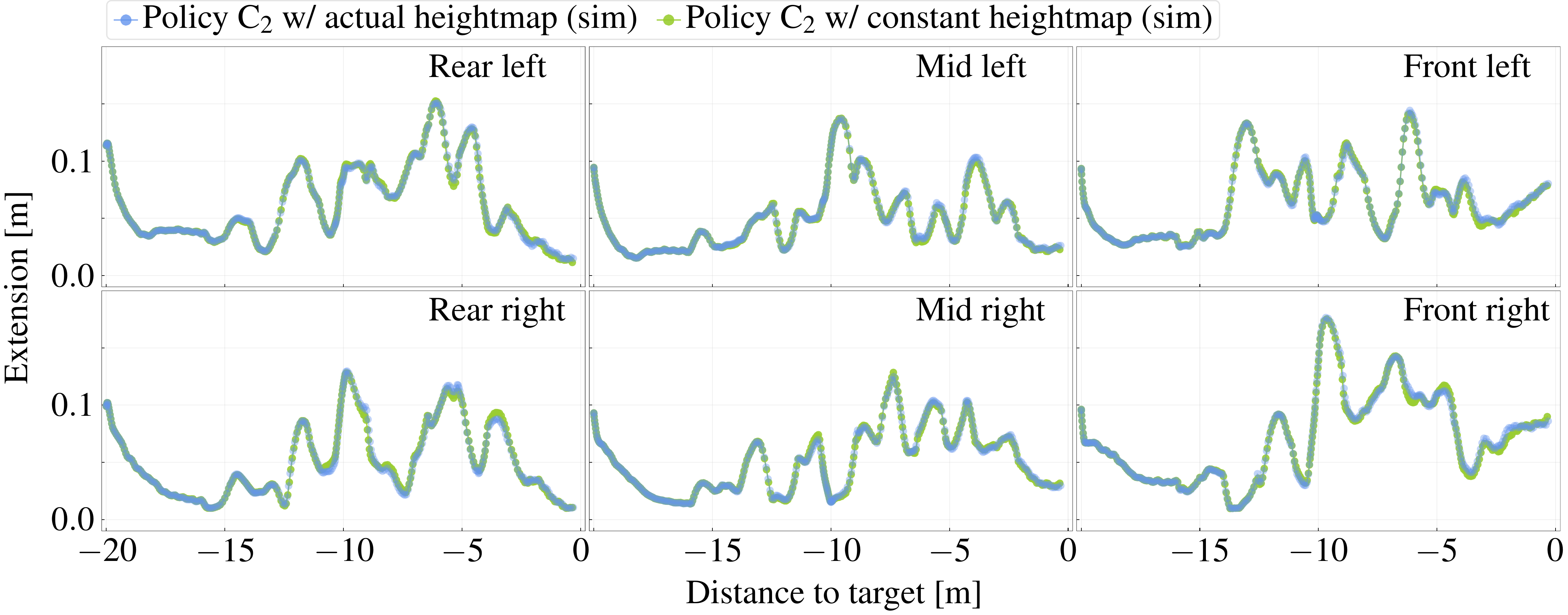}
    \caption{Comparison between two simulation runs over the vibration course, one with obstacles included in the height map and one without.}
    \label{fig:vibration_course_sim_v_sim}
\end{figure}

Instead, we resort to simulation, motivated by the idea that if the controller does not use the height map there, it will not do so in reality either.
In the simulation, we embed obstacles in the surface model and conduct two experiments identical to the field experiments, except we use a completely flat ground to avoid noisy measurements, see Fig.~\ref{fig:vibration_course_sim_v_sim}.
The information from the height map does not affect the arms' extension, so we conclude that the controller mainly uses proprioceptive information for decision-making.
We did note that, for other constant height maps of extreme values, the controller gets stuck or performs poorly, indicating that the local heights are of some use.
The use is, however, not for predictive planning, as we had hoped and anticipated from previous work~\cite{wiberg2021control}.



%

\begin{figure}
    \centering
    \includegraphics[width=0.7\columnwidth]{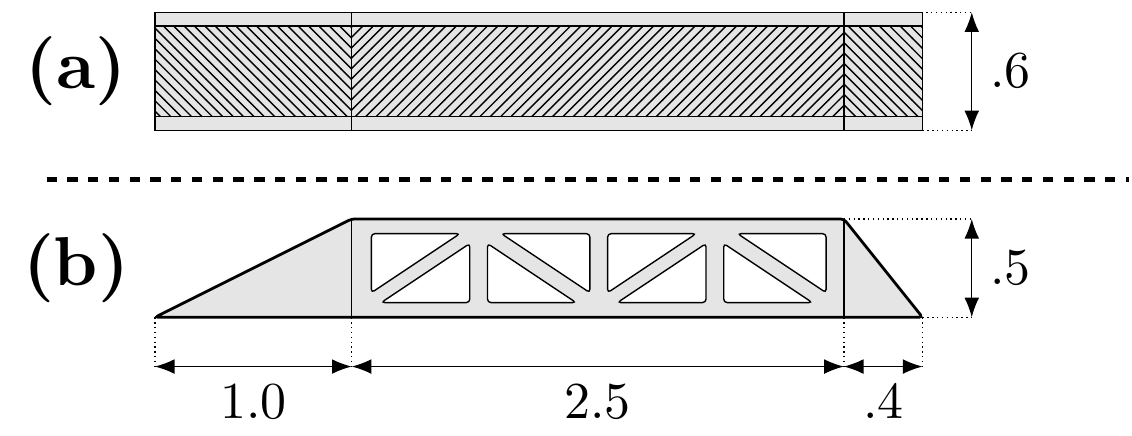}
    \caption{Schematic of the ramp obstacle from a top-down view (a) and side view (b). The vehicle drives from left to right in the figure. All values are in metres.}
    \label{fig:ramps_schematic}
\end{figure}

\subsection{Obstacles}
To study the use of the suspensions in isolation, we use a ramp wide enough to fit one wheel, see Fig.~\ref{fig:ramps_schematic}.
For this case, the controller uses Policy~$\mathrm{C}_2$, which has trained on terrains with boulders, including observation noise and action delays.
The primary objective of this experiment is to evaluate the sim-to-real transfer capability, with a secondary focus on examining the learnt control policy.

\begin{figure}
    \centering
    \includegraphics[width=\columnwidth]{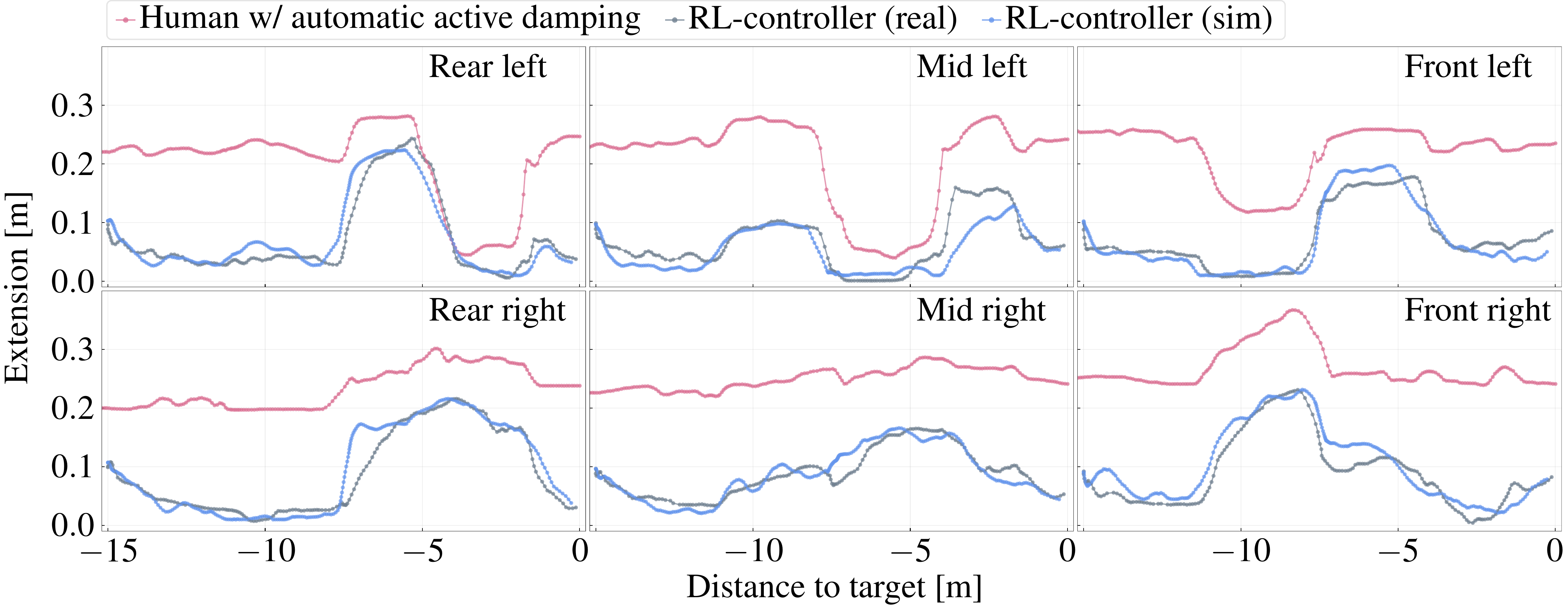}
    \caption{Arm extension for all wheels when driving over the large ramp. The real and simulation case uses Policy~$\mathrm{C}_2$. In the human case, the suspensions were regulated by a conventional controller that has a nominal extension of 0.25~m and twice the range of the DRL controllers.}
    \label{fig:ramps_arm_pos}
\end{figure}

The DRL controller successfully controls the arms to overcome the big ramp and reach the target placed 15~m straight ahead.
For a comparison with the simulation, we use the 3D representation of the scanned gravel parking area with embedded obstacles and place the vehicle and target at the same coordinates as during the real experiment, see Fig.~\ref{fig:photo_series}.
A sim-vs-real video of this scenario is available in the supplementary material.
When we study the arm extensions, it is clear that the policy developed in simulation to control the suspensions works equally well in reality, see Fig.~\ref{fig:ramps_arm_pos}.
As the front left wheel moves up the ramp, the controller tries to lift it while pressing down on the other side to maintain levelling and balance ground forces. 
We see similar behaviour for the wheels on the middle and rear sections.
With locked suspensions, the vehicle's maximum roll angle would be roughly 14 degrees, which is reduced to 6 degrees with the DRL control.

\begin{figure*}
    \centering
    \includegraphics[clip, trim=0.1cm 0.0cm 0.0cm 2.0cm, width=\textwidth]
    {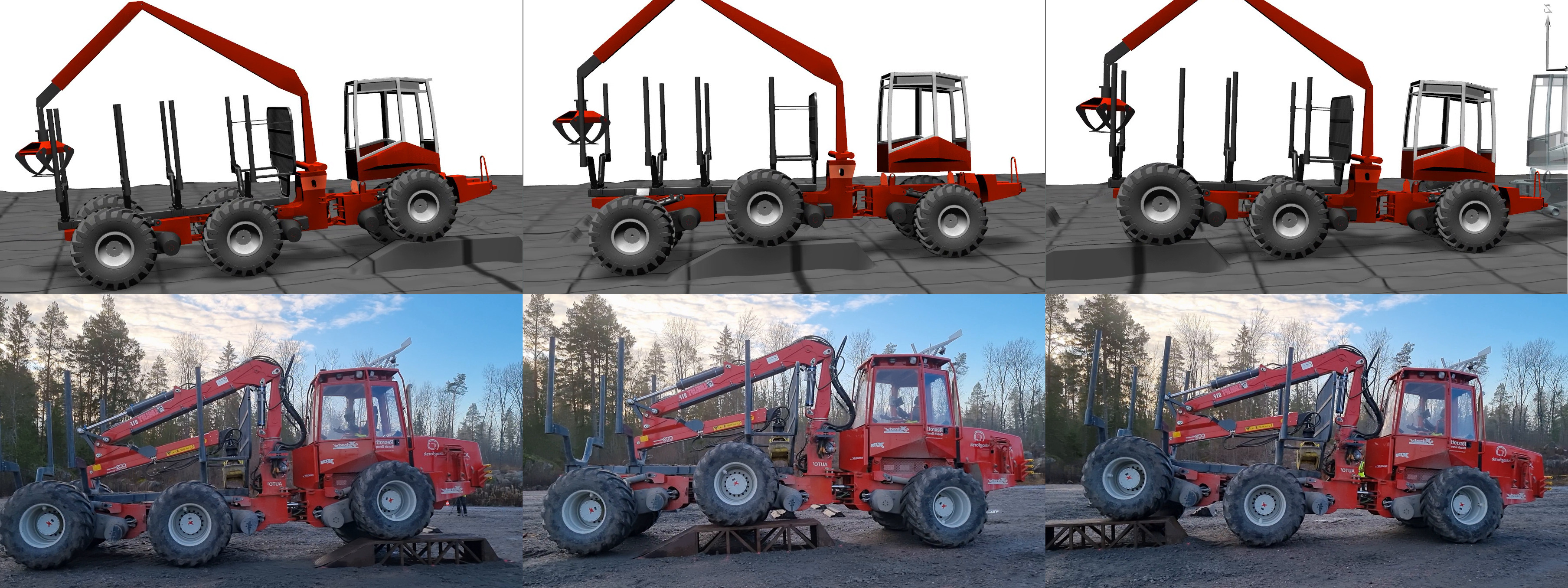}
    \caption{Sim-vs-real comparison of Policy~$\mathrm{C}_2$ driving over the large ramp. Images are mirrored to run from left to right.}
    \label{fig:photo_series}
\end{figure*}

As seen in Fig.~\ref{fig:ramps_arm_pos}, we also compare the DRL controller to
the automatic suspension control with a human operator for steering and throttle.
The trends in arm extensions show a clear resemblance, indicating that the DRL controller has learnt a similar policy to balance ground forces.
The human case results in a maximum roll angle of 4 degrees, although the comparison is not entirely fair.
For reasons of safety and comfort, the human operator drove significantly slower than the DRL controller, giving more time for the actuators to level out, and was allowed to use twice the suspension range.
When placing the ramp on the other side of the vehicle, the DRL controller got stuck on the chassis.
However, considering the small margins with the limited suspension range, this is a challenging scenario.

\begin{figure}
    \centering
    \includegraphics[width=\columnwidth]{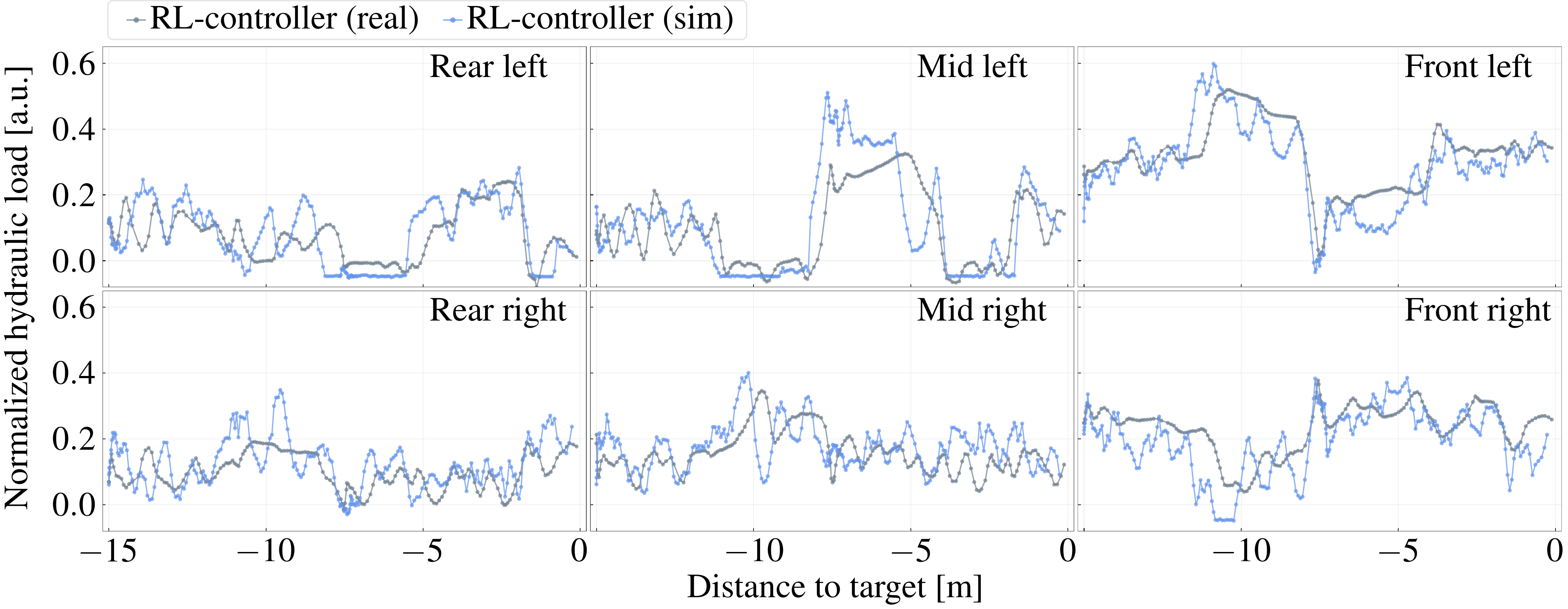}
    \caption{Hydraulic load on the suspensions when driving over the large ramp, where both the real and the simulation case use Policy~$\mathrm{C}_2$.}
    \label{fig:ramps_arm_force}
\end{figure}

Having previously established that the local height map is insignificant in controlling suspensions, the hydraulic load must act as the primary feedback.
However, since the system identification in Section~\ref{sec:sys:arms} uses calibration scenarios on level ground, it is not clear how well it applies to the ramp experiments.
Fig.~\ref{fig:ramps_arm_force} compares the normalized hydraulic loads when passing the ramp, which show qualitative agreement, indicating that the system identification has captured the dynamics beyond the calibration scenarios.
The main difference between the curves is that the simulation shows larger fluctuations.
This is because the simulation model assumes rigid wheels and compliant lock constraints for the suspensions, whereas the real system has flexible tyres and hydraulic accumulators that provide additional smoothing and damping.
While these differences do not seem to affect performance, they imply the need for a tyre model to increase simulation fidelity.
Interestingly, the ``noise'' in simulated hydraulic load has the same effect as domain randomization.
This explains why the policies are so robust to the domain shift and perform better than expected in the real world despite not adding noise to the observations for the hydraulic load.

\section{Conclusion}
We conclude that an accurate model of the actuator dynamics, modelling action delays, and preventing the emergence of bang-bang control are necessary to handle the sim-to-real gap for heavy vehicles with hydraulic actuators in rough terrain.

Actuator models using 1D motor constraints and system identification produce sufficiently accurate models of the hydraulic actuators.
This approach is faster to simulate and requires fewer parameters for tuning than modelling the actual hydraulic circuits.
Moreover, some of the success is attributed to the low-level PID controllers, which operate independently of the DRL controller and are responsible for the final actuation of the system.
Tuning the PIDs in simulations allows slack in modelling errors and can mitigate the effect of unmodelled actuator dynamics.
A higher fidelity simulation would involve a tyre model but does not appear to be necessary for the studied control task.
Natural terrain may amplify the effects of the simplified tyre model.
However, due to wet November conditions, we were unable to transport the vehicle to such areas.

Aside from the actuator model, including actuator delays during training have the most prominent effect on sim-to-real transfer.
The impact of adding observation noise is unclear since the simulated ground forces, the primary feedback for suspension control, are already noisy.
As a result, the effect on controller robustness is similar to domain randomization.

The controllers we present have room for improvement when it comes to obstacles and making use of perceptual information.
Due to slow learning, the current study did not reach the level of learning to distinguish between passable and unpassable obstacles as in our previous work~\cite{wiberg2021control}.
This is partly attributed to the vehicle model of the hydrostatic transmission with two exotic differential gears.
Compared to the previous study, which employed individual wheel torque control, the current transmission presents a more complex state-action-reward mapping, necessitating further effort.
Another potential reason is the limited available suspension range and speed enforced for safety reasons, which restricts exploration.
To address these complexities, a promising approach is to take inspiration from similar work in legged locomotion, e.g., by using an attention-based recurrent encoder~\cite{miki2022learning}.
These encoders could simplify learning about the transmission and enhance the use of the suspensions by coupling proprioceptive and exteroceptive information in predictive planning.
On the upside, we now know that the control policies learnt in simulation transfer to reality.

\section{Acknowledgements}
The research was supported in part by Troëdsson Teleoperation Lab, Mistra Digital Forest, Algoryx Simulation AB, Swedish National Infrastructure for Computing at High-Performance Computing Center North (HPC2N), eSSENCE, and eXtractor AB.

\section{Declaration of generative AI and AI-assisted technologies in the writing process}
During the preparation of this work, the authors used ChatGPT 3.5 to enhance text clarity, fluency, and conciseness.
After using this tool, the authors reviewed and edited the content as needed and take full responsibility for the content of the publication.

\bibliographystyle{elsarticle-num}
\bibliography{bib}


\end{document}